\documentclass[10pt,onecolumn]{article}

\usepackage[a4paper,margin=0.8in]{geometry}

\usepackage{amsmath}
\usepackage{amssymb}
\usepackage{mathtools}
\usepackage{bm}

\usepackage{graphicx}
\usepackage{booktabs}
\usepackage{multirow}
\usepackage{siunitx}
\usepackage{adjustbox}
\usepackage{xcolor}
\usepackage{wrapfig}
\usepackage{placeins}
\usepackage{subcaption}
\usepackage{makecell}

\usepackage{comment}
\usepackage{authblk}
\usepackage{orcidlink}
\usepackage[numbers,sort&compress]{natbib}

\title{
    \textbf{Continuous-Time Cloth Dynamics
    with Physics-Informed Neural ODEs}
}

\author[1]{Yingdong Ru\textsuperscript{*}}
\author[1]{Lipeng Zhuang\textsuperscript{*}}
\author[1]{Zhuo He}
\author[1]{Gerardo Aragon Camarasa}

\affil[1]{
    University of Glasgow, Glasgow, United Kingdom
}

\date{}

\begin{document}

\maketitle

\begingroup
\renewcommand{\thefootnote}{\fnsymbol{footnote}}
\footnotetext[1]{Equal contribution.}
\endgroup

\begin{abstract}

We present ContinuousCloth, a \textit{continuous-time, physics-informed} neural solver for deformable cloth dynamics that enforces a spring--mass ODE within a neural trajectory model. Unlike classical time-stepping simulators that return trajectories only on a discrete integration grid, ContinuousCloth supports querying at arbitrary timestamps and computing displacement, velocity, and acceleration via automatic differentiation. We evaluate ContinuousCloth along two dimensions. In a physics-only setting, ContinuousCloth achieves trajectory accuracy comparable to high-precision solvers across six diverse boundary-condition scenarios while reducing runtime by up to $17.4\times$. When fused with sparse and noisy observations, the spring--mass ODE acts as a strong inductive bias for downstream tasks: interpolation and parameter estimation. That is, ContinuousCloth reduces interpolation error by $7.3\times$ over a data-only neural baseline in simulation and $2.4\times$ on real-world cloth sequences, and achieves $2.6\times$ lower parameter error than a differentiable-solver baseline under high-noise conditions, demonstrating consistent robustness under realistic measurement imperfections.

\end{abstract}

\label{sec:intro}

\section{Introduction}

\begin{figure*}[t]
    \center
    \includegraphics[width=0.90\linewidth]{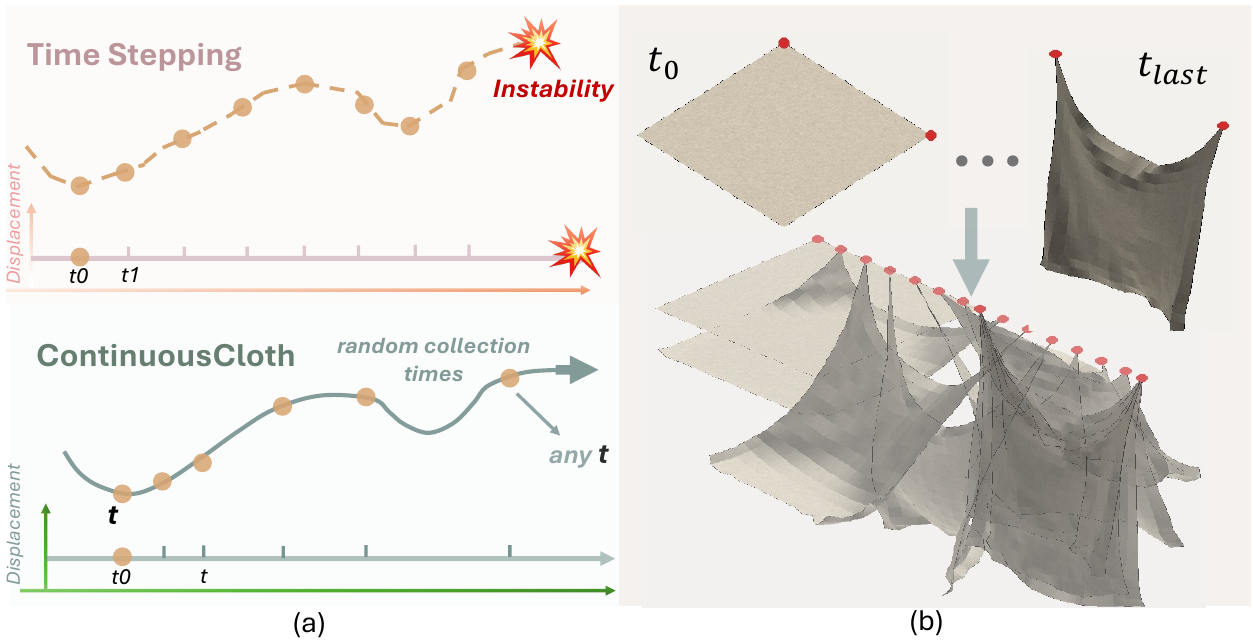}
    \vspace{4mm}
    \caption{(a) Classic simulators accumulate errors on discrete 
grids and can become unstable, whereas ContinuousCloth represents 
cloth motion as a continuous function queryable at any $t$. 
(b) Given only the first and last frames, ContinuousCloth 
reconstructs a physically consistent trajectory across the full 
intermediate horizon.}
    \label{fig:teaser}
\end{figure*}

Classic cloth simulators discretize motion into time steps, returning states only on a solver-defined time grid \cite{qiao2020scalable,hudifftaichi,li2022diffcloth}. This discrete time-stepping introduces several practical limitations. First, rollout accuracy and stability are governed by the chosen step size. That is, coarse time steps accumulate local numerical errors over long horizons and can cause numerical instability, leading the simulation to diverge and producing physically inaccurate cloth motions in stiff regimes, while reducing the step size increases computational cost proportionally\cite{li2020p}. Second, evaluating motion at unseen timesteps, such as irregular observation times or a different time stepping rate, requires either re-running the solver or relying on discretization-dependent interpolation\cite{he2025physically,breen2000survey}. Third, derivative quantities such as velocity and acceleration are typically obtained via finite differences, making them sensitive to the chosen timestep and noise\cite{li2023review}.

Recent progress in Physics-Informed Neural Networks (PINNs)\cite{lawal2022physics} and the demonstrated effectiveness of neural fields motivate us to ask whether \textit{continuous, coordinate-based representations can help overcome these limitations}. PINNs offer a complementary alternative to time stepping by learning the solution as an explicit, differentiable function and enforcing the governing equations through residual minimization\cite{bararnia2022application}. However, existing physics-informed neural solvers have primarily focused on low-dimensional deformable linear objects, or quasi-static fabric deformation, while continuous-time modeling of high-dimensional cloth dynamics remains underexplored.

\begin{figure*}[t]
    \center
    \includegraphics[width=0.9\linewidth]{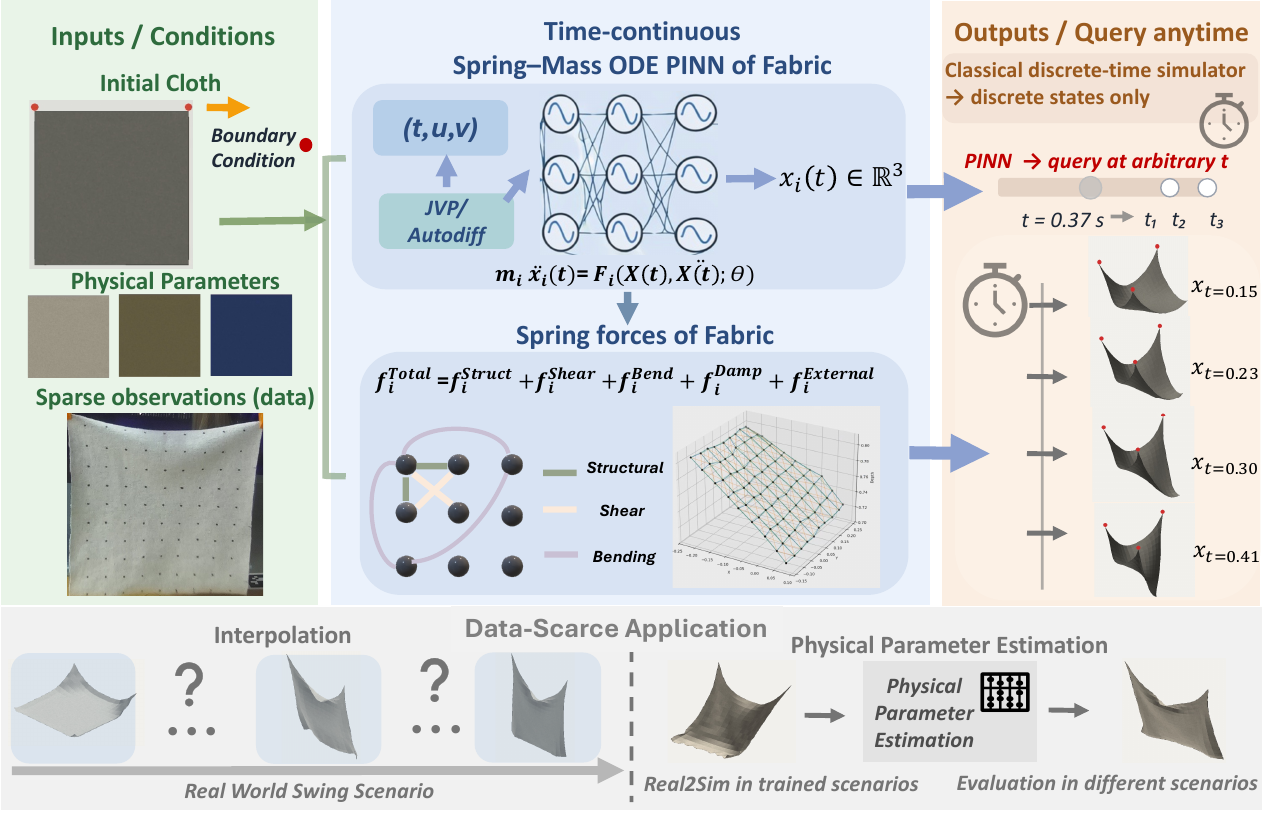}
    \vspace{4mm}
    \caption{Overview of ContinuousCloth. We represent cloth dynamics as a continuous-time trajectory field conditioned on boundary conditions and physical parameters, and enforce spring--mass ODE constraints via physics residuals computed with automatic differentiation. ContinuousCloth can be queried at arbitrary timestamps and the same framework supports data-scarce applications, including trajectory interpolation from sparse observations and inverse physical-parameter estimation.}
    \label{fig:main}
\end{figure*}

Building on this idea, we propose \textbf{{ContinuousCloth}} (Fig. \ref{fig:main}), a continuous-time physics-informed neural solver for deformable cloth dynamics. Although kirchhoff-love thin-shell PDEs\cite{ciariet2000mathematical} provide more physically accurate cloth behaviors, solving them as residuals using a neural network for full continuous-time cloth trajectories is difficult in practice. This difficulty is not only caused by the fourth-order spatial derivatives involved in strong-form thin-shell residuals, but also by the ill-conditioned space-time optimization required to learn dynamic cloth trajectories over both space and time, even under energy-based formulations\cite{kairanda2024neuralclothsim}. Moreover, sparse collocation of either space and time can miss deformation details of fabrics in continuous-time trajectories\cite{bastek2023physics}, whereas dense collocation greatly increases automatic-differentiation cost and training time\cite{cho2023separable}. We therefore decide to use the spring--mass ODE that balances fidelity and learnability. It captures the dominant elastic and damping structure of cloth motion while remaining efficient, differentiable, continuous time and identifiable physical prior. ContinuousCloth represents cloth motion as a continuous-time trajectory field which is constrained by a spring–mass ODE (see Fig.\ref{fig:teaser} for an overview). Given initial and boundary conditions, the model is trained by minimizing physics residuals over the continuous time horizon.
Positions are obtained directly from the trajectory field, while velocities, accelerations, and forces are computed through automatic differentiation and physics equations. This representation decouples trajectory evaluation from a fixed integration step size and, once trained, the solver can be queried at arbitrary timestamps without re-running an integrator, while remaining globally constrained by the underlying dynamics over the full time horizon. The formulation is architecture-agnostic, and we instantiate it with both MLP (ContinuousCloth-MLP) and transformer (ContinuousCloth-Former) backbones in our experiments. However, learning continuous-time dynamics for high-dimensional cloth is challenging, since the model must capture both rapid local deformation and long-horizon oscillatory behavior over a large coupled particle system. To address continuous-time learning, we adopt a \emph{multi-resolution Fourier pseudo-sequence} training strategy. It combines spatiotemporal Fourier features with variable-length temporal windows, exposing the model to both short-horizon deformation and long-horizon oscillatory dynamics.

ContinuousCloth can be used in two regimes: a physics-only setting and data-assisted (i.e. data-scarce) settings. In the physics-only setting, it serves as a data-free forward solver for dynamic spring–mass cloth. This setting tests whether a continuous-time neural field can faithfully reproduce cloth dynamics under diverse boundary conditions and physical parameters. Across six boundary-condition scenarios, ContinuousCloth achieves trajectory accuracies comparable to a high-precision fourth-order Runge--Kutta (RK4) reference\cite{yang2013cloth} while reducing dense-time evaluation runtime by up to $17.4\times$. A single conditioned model also generalizes across different boundary conditions and material parameters of fabrics, achieving low residuals and constant accuracy results.

In data-assisted settings where real cloth observations are sparse, irregular, and noisy, purely data-assisted models can fit such measurements but frequently lack temporal consistency and overfit measurement noise. ContinuousCloth addresses this by coupling data fitting with physics-based regularization in a single objective, where the ODE residual serves as an inductive bias that guides the solution toward physically plausible motion between observations. As a result, the same formulation supports practical downstream tasks, including trajectory interpolation and inverse parameter estimation. We validate these capabilities in simulation under controlled sparsity and noise 
and on \emph{real-world cloth sequences}, where ContinuousCloth-MLP reduces 
interpolation error by $7.3\times$ on average over a data-only neural baseline 
and achieves up to $2.6\times$ lower physical parameter error than the 
differentiable solver baseline under large-noise conditions in simulation, demonstrating 
consistent improvements in robustness under realistic measurement imperfections.


In summary, our main contributions are:
\begin{itemize}
    \item We present ContinuousCloth, a PINN-based \emph{continuous-time} neural solver for spring-mass cloth dynamics that represents the cloth trajectory as an explicit differentiable function of time and enforces the governing ODE via physics-residual minimization instead of discrete time stepping and finite differences.
    \item We formulate ContinuousCloth as a unified continuous-time framework for both data-free and data-scarce settings. 
    In the physics-only, data-free method, ContinuousCloth matches high-precision solver accuracies while enabling faster dense-time evaluation; while in the data-scarce with sparse and noisy observations, the same full-horizon ODE regularization serves as a physical prior for consistent accurate interpolation and noise-robust inverse parameter estimation.
    \item We propose a multi-resolution pseudo-sequence generator for training ContinuousCloth, which constructs variable-length temporal windows with a fixed number of temporal collocation points. This provides learning of both rapid local deformations and long-horizon oscillatory fabric dynamics.

\end{itemize}


\section{Literature Review}
\paragraph{PINNs for deformable objects:}

Deformable object manipulation remains a central problem in robotics, where both Model Predictive Control and Reinforcement learning methods depend on cloth simulation~\cite{longhini2025unfolding,zhang2025particle}. Existing simulators face a trade-off between physical accuracy and computational efficiency, while data-scarce neural emulators require large, densely labeled datasets and are sensitive to noise in real-world observations~\cite{karamangil2010simulation,zhang2025dynamic}. PINNs provide an alternative by embedding physical laws into the model architecture, allowing physics-consistent inference from sparse supervision~\cite{xia2024physics}. Specifically, PINNs have been shown to solve a wide range of complex physical PDEs, such as Navier–Stokes equations \cite{zhouhybrid,ali2025machine,zhang2025pseudo,li2025pinn,khademi2025physics} and poisson equations \cite{wu2023physics,hu2024solving,zhang2023physics,zhongkai2024pinnacle,kamali2023elasticity}.

Several studies have also applied PINNs to model the behavior of deformable objects\cite{xu2023transfer}. For instance, Zhou el al.\cite{zhou2023transfer} use PINNs with transfer learning to address parameter estimation in metal plates and beams, particularly when data is scarce and costly to obtain. Another example is Rao et al.~\cite{rao2021physics}, who addresses the problem of wave propagation and 2d-defected plate under forces using PINNs. NeuralClothSim~\cite{kairanda2024neuralclothsim} models cloth as a continuous neural field within the \textit{Kirchhoff–Love} thin-shell framework. It yields a resolution-independent, differentiable representation that can be queried at arbitrary spatial points, offering an alternative to mesh-dependent simulators. However, its emphasis is on quasi-static equilibrium rather than continuous, time-dependent cloth dynamics.

In parallel, several works have validated PINNs as continuous-time solvers for low-dimensional dynamical systems. Martin et al. \cite{muzelak2024physics} introduces the use of PINNs to solve the dynamics of a spring using a spring-mass ODE. 
Sahoo et al. \cite{sahoo2024application} apply PINNs to the dynamics of the mass-spring-damper and study PINNs as forward solvers on several typical vibration benchmarks. Their learned model approximates the displacement response under representative settings such as damped free decay and forced oscillations with and without damping. Eid et al. \cite{eid2025dynamic} further studies a coupled dual-spring pendulum and use a PINN to learn the system’s nonlinear oscillatory dynamics. The results show that the learned network preserves phase-space structure, achieves high agreement with reference trajectories across multiple initial conditions, and remains robust even in strongly nonlinear regimes. 

Prior work has demonstrated the effectiveness of PINNs on continuous-time deformable linear objects dynamics  \cite{muzelak2024physics,eid2025dynamic,rao2021physics} and final state equilibrium deformation of fabrics. In contrast, continuous-time fabric dynamics remain less explored in settings where deformation is high-dimensional, motivating our focus on modeling spring-mass cloth dynamics as a continuous-time PINNs solver. 

\vspace{-2mm}
\paragraph{Deep learning combined with physics guidance:} Recent years have seen growing interest in combining deep learning with physical guidance for deformable object modeling and manipulation. DEFORM \cite{chen2025differentiable} integrates differentiable discrete elastic rod dynamics with learning to achieve real-time modeling, tracking, planning and control for deformable linear objects. PhysTwin \cite{jiang2025phystwin} shows that sparse videos can be converted into digital twins by jointly optimizing geometry, appearance, and physical parameters under a physics-informed representation, bridging visual perception, downstream simulation and motion planning tasks. Similarly, robotic control-oriented physics-informed learning has also emerged for deformable linear objects. For instance, Du et al. \cite{du2025physics} proposes a Cosserat-based PINN for real-time deformable linear objects modeling and MPC-based shape control, highlighting the potential of PINNs for sample-efficient and physics-consistent manipulation. Although recent work has demonstrated the value of physics-guided learning for deformable objects, most existing methods focus either on control-oriented deformable linear objects systems (e.g., rods, ropes, cables) or on digital-twin reconstruction from videos, rather than continuous-time and continuous-space cloth dynamics modeling with sparse observations.

\section{Method}
As illustrated in Figure~\ref{fig:main}, we model fabric dynamics as a spring-mass ODE system over a regular vertex grid, and parameterize the continuous-time solution as a neural field mapping of spatiotemporal coordinates to 3D vertex positions. Time derivatives required by the ODE are computed via automatic differentiation, and the network is trained by jointly minimizing physics residuals, boundary, initial condition, and optional data losses when sparse observations are available. Unlike classical discrete-time simulators that produce states only at fixed time steps, the trained network supports queries at arbitrary time, and naturally extends to data-scarce applications such as temporal interpolation and physical parameter estimation as shown in Sec. \ref{sec:experiment}.


Specifically, we model a deformable fabric as a regular spring grid with $n_x \times n_y$ vertices.
Let $\mathcal{V}=\{1,\dots,N\}$ be the vertex set, where $N=n_x n_y$ is the total number of vertices.
Each vertex $i$ is associated with normalized material coordinates
$\mathbf{s}_i=(u_i,v_i)\in[0,1]^2$ and has 3D position $\mathbf{x}_i(t)\in\mathbb{R}^3$ at \allowbreak\ time $t\in[0,T]$. We represent the continuous-time cloth motion using an implicit neural field 
\begin{equation}
\mathbf{x}_i(t)=\mathbf{X}_\phi(t,\mathbf{s}_i),\qquad
\mathbf{X}_\phi:[0,T]\times[0,1]^2\rightarrow\mathbb{R}^3,
\end{equation}
parameterized by $\phi$.
Time derivatives $\dot{\mathbf{x}}_i(t)$ and $\ddot{\mathbf{x}}_i(t)$ are obtained by differentiating
$\mathbf{X}_\phi$ w.r.t.\ $t$.

\vspace{2mm}
\noindent\textbf{Boundary and initial conditions:} We split vertices into fixed boundary, $\mathcal{V}_{\text{fix}}$,
moving boundary, $\mathcal{V}_{\text{mov}}$, and free vertices, 
$\mathcal{V}_{\text{free}}=\mathcal{V}\setminus(\mathcal{V}_{\text{fix}}\cup\mathcal{V}_{\text{mov}})$.
For boundary vertices $i\in\mathcal{V}_{\text{fix}}\cup\mathcal{V}_{\text{mov}}$, the target trajectories are known. The initial state of $(\mathbf{x}_i^0,\mathbf{v}_i^0)$ are also available for all vertices.

\vspace{2mm}
\noindent\textbf{Data fusion for downstream tasks:}
If discrete observations are available at frame times $\{t_k\}_{k=1}^K$,
we denote them as $\tilde{\mathbf{x}}_i(t_k)$.  Our goal is to learn $\mathbf{X}_\phi$ while satisfying boundary, initial and physics constraints, and matching observations when provided.

\subsection{Spring--Mass Cloth Dynamics}

We use a standard spring-mass cloth model with three spring types:
(i) \emph{structural} springs connecting neighborhood grid edges,
(ii) \emph{shear} springs connecting diagonal neighbors, and
(iii) \emph{bending} springs connecting 2-hop neighbors along grid axes.  Let a vertex be indexed by grid coordinate $(p,q)$ with $p\in\{1,\dots,n_x\}$ and $q\in\{1,\dots,n_y\}$ (Fig. \ref{fig:main}.
Therefore, we define undirected edge sets as:
\begin{align}
\mathcal{E}_{\text{structural}} &= \{((p,q),(p{+}1,q)),\ ((p,q),(p,q{+}1))\},\\
\mathcal{E}_{\text{shear}} &= \{((p,q),(p{+}1,q{+}1)),\ ((p,q),(p{+}1,q{-}1))\},\\
\mathcal{E}_{\text{bending}} &= \{((p,q),(p{+}2,q)),\ ((p,q),(p,q{+}2)), ((p,q),(p{+}2,q{+}2))\},
\end{align}
where edges that fall outside the grid are omitted (in implementation, these pairs map to vertex ids in $\mathcal{V}$.) Each vertex $i$ has mass $m_i$.

For an edge $(i,j)\in\mathcal{E}_r$ of type $r\in\{\text{\textit{structural}},\text{\textit{shear}},\text{\textit{bending}}\}$,
let $l_{ij}^0$ be its rest length of the spring (e.g., from the rest/initial configuration), we define
\begin{equation}
\mathbf{d}_{ij}=\mathbf{x}_i-\mathbf{x}_j,\quad
l_{ij}=\|\mathbf{d}_{ij}\|_2,\quad
\hat{\mathbf{d}}_{ij}=\mathbf{d}_{ij}/(l_{ij}+\epsilon),
\end{equation} as the edge direction ($\mathbf{d}_{ij}$), edge length ($l_{ij}$), derivative of edge direction ($\hat{\mathbf{d}}_{ij}$) respectively. $\epsilon$ is a small number for numerical stability.
We use an elastic spring force ($\mathbf{f}^\text{\textit{elas}}$) with damping ($\mathbf{f}^\text{\textit{damp}}$) along the spring direction:
\begin{align}
\mathbf{f}^{\text{\textit{elas}}}_{ij} &= -k_r (l_{ij}-l_{ij}^0)\,\hat{\mathbf{d}}_{ij}, \\
\mathbf{f}^{\text{\textit{damp}}}_{ij} &= -c_r \big((\dot{\mathbf{x}}_i-\dot{\mathbf{x}}_j)^\top \hat{\mathbf{d}}_{ij}\big)\,\hat{\mathbf{d}}_{ij},
\end{align}
where $(k_r,c_r)$ are stiffness and damping coefficients for spring type $r$. The total internal force on vertex $i$ is the sum over its incident springs:
\begin{equation}
\mathbf{f}^{\text{\textit{int}}}_i =
\sum_{r}\sum_{j:(i,j)\in\mathcal{E}_r}\left(\mathbf{f}^{\text{elas}}_{ij}+\mathbf{f}^{\text{damp}}_{ij}\right).
\end{equation}

We incorporate gravity $\mathbf{f}^{\text{\textit{grav}}}_i=m_i\mathbf{g}$ and an optional global drag force
$\mathbf{f}^{\text{\textit{drag}}}_i=-c_g\dot{\mathbf{x}}_i$ as the external force where $\mathbf{g}$ is the acceleration of gravity, $c_g$ is the global damping coefficient which affects how quickly the system changes under external forces.
Let $\mathbf{F}_i(\mathbf{X},\dot{\mathbf{X}};\theta)=\mathbf{f}^{\text{\textit{int}}}_i+\mathbf{f}^{\text{\textit{grav}}}_i+\mathbf{f}^{\text{\textit{drag}}}_i$,
with $\theta=\{k_r,c_r,c_g\}$.
For free vertices, the dynamics follow the second-order ODE:
\begin{equation}
m_i\ddot{\mathbf{x}}_i(t) = \mathbf{F}_i(\mathbf{X}(t),\dot{\mathbf{X}}(t);\theta),
\qquad i\in\mathcal{V}_{\text{free}}.
\end{equation}

\subsection{Continuous-time neural representation}
Our physics-informed formulation is \textit{model-agnostic} in this work. We evaluate two network architectural variants:  MLP-based and Transformer-based architectures. Detailed architectural configurations and hyper-parameters for each variant are provided in the Supplementary file.

To model the continuous spatial-temporal dynamics and long-term dependencies in sequential data, we propose a multi-resolution training scheme with Fourier feature embedding to capture the position variation of the spring-mass grid against conventional PINNs on fixed sequential data.
We first transform point-wise spatiotemporal inputs into temporal sequence following the pseudo sequence generator (PSG)~\cite{zhao2023pinnsformer}, then incorporating multi-scale Fourier feature embeddings\cite{wang2021eigenvector} to enhance dynamical feature modeling during fabric simulation.
Specifically, the spatiotemporal coordinates $s,t$ are encoded through Fourier transformation with multiple frequencies and then these embeddings are concatenated together as the high-dimensional Fourier representation (See Eqn.~\ref{eqn:posemb}).
\begin{equation}
    \begin{gathered}
        \mathcal{B}_s=\{2^0,...,2^{F_s-1}\}, \mathcal{B}_t=\{2^0,...,2^{F_t-1}\},\\
        \mathcal{F}(s,t)=[s,t,\{sin(2\pi\beta\mathrm{x}),cos(2\pi\beta\mathrm{x})\}_{\beta\in\mathcal{B}_s},\{sin(2\pi\beta\mathrm{x}),cos(2\pi\beta\mathrm{x})\}_{\beta\in\mathcal{B}_t}]
        \label{eqn:posemb}
    \end{gathered}
\end{equation}
where $F_s, F_t$ are the total frequency amount of time and position coordinates, respectively, and $\beta$~denotes frequency bands sampled from $\mathcal{B}_s$ and $\mathcal{B}_t$, respectively, \allowbreak\ enabling sine/cosine encodings at exponentially increasing frequencies. $[\cdot]$ indicates the concatenation operation. 

Furthermore, we propose a sub-sequence resampling strategy for the embedded features to improve the dynamical modeling capacity as the extension of PSG, named \textit{Multi-resolution Fourier Pseudo Sequence Generator} (MFPSG). Specifically, the sub-sequence of time $T_{sub}$ is a variable-length temporal window defined by start and end time points ($t_a, t_b$) within the whole time range $T_{total}$, Each sub-window can have a different duration, but is uniformly sampled into a fixed number of temporal collocation points. During training, mini-batches are constructed by sampling multiple pseudo-sequences, allowing the model to learn from temporal windows at different resolutions. (See Eqn.~\ref{eqn:mfpsg}).
    
\begin{equation}
    \label{eqn:mfpsg}
    \begin{gathered}
        T_{\text{sub}}=[t_a,t_b]\subseteq T_{\text{total}}, \quad t_a<t_b, \quad k\in\mathbb{N}.\\
        \Delta t=\frac{t_b-t_a}{k-1}, \quad t_i=t_a+(i-1)\Delta t,\ i=1,\dots,k.\\
        \mathcal{X}(s;T_{\text{sub}})=\big[\mathcal{F}(s,t_1),\dots,\mathcal{F}(s,t_k)\big].
    \end{gathered}
\end{equation}
where $\Delta t$ is the minimized time span of each training epoch, $\mathcal{F}$ is the Fourier feature embedding introduced in the first stage. The varying time range with fixed time sample points produces a multi-resolution spatial-temporal sequence $\mathcal{X}(s;T_{\text{sub}})$ which facilitates training PINNs as shown in Tab.\ref{tab:scenario_compare}.


\subsection{Physics-Informed Objective and Constraints}
We optimize $\phi$ (and optionally $\theta$ for physics parameter estimation) by minimizing physics residuals and constraint losses.
For each $i\in\mathcal{V}_{\text{free}}$, we define the dynamics residual as
\begin{equation}
\mathbf{r}_i(t)= m_i\ddot{\mathbf{x}}_i(t) - \mathbf{F}_i(\mathbf{X}(t),\dot{\mathbf{X}}(t);\theta).
\end{equation}
For the physics residual loss, we have the following given collocation times $\mathcal{T}_{\text{phys}}\subset[0,T]$,
\begin{equation}
\mathcal{L}_{\text{phys}} =
\mathbb{E}_{t\sim\mathcal{T}_{\text{phys}}}\left[\frac{1}{|\mathcal{V}_{\text{free}}|}
\sum_{i\in\mathcal{V}_{\text{free}}}\|\mathbf{r}_i(t)\|_2^2\right].
\end{equation}
For boundary vertices $\mathcal{V}_{\text{bc}}=\mathcal{V}_{\text{fix}}\cup\mathcal{V}_{\text{mov}}$ (boundary constrains) with targets
$\bar{\mathbf{x}}_i(t)$ and enforcement times $\mathcal{T}_{\text{bc}}$,
we impose a soft penalty as follows:
\begin{equation}
\mathcal{L}_{\text{bc}} =
\mathbb{E}_{t\sim\mathcal{T}_{\text{bc}}}\left[\frac{1}{|\mathcal{V}_{\text{bc}}|}
\sum_{i\in\mathcal{V}_{\text{bc}}}\|\mathbf{x}_i(t)-\bar{\mathbf{x}}_i(t)\|_2^2\right],
\end{equation}
For the initial conditions, we enforce the initial state at $t=0$ by:
\begin{equation}
\mathcal{L}_{\text{ic}}=
\frac{1}{|\mathcal{V}|}\sum_{i\in\mathcal{V}}
\left(\|\mathbf{x}_i(0)-\mathbf{x}_i^0\|_2^2 + \|\dot{\mathbf{x}}_i(0)-\mathbf{v}_i^0\|_2^2\right).
\end{equation}
Last, when data observations, $\tilde{\mathbf{x}}_i(t_k)$, are available, we have:
\begin{equation}
\mathcal{L}_{\text{data}}=
\frac{1}{\sum_{k,i} M_{k,i}}
\sum_{k=1}^K\sum_{i\in\mathcal{V}}
M_{k,i}\,\|\mathbf{x}_i(t_k)-\tilde{\mathbf{x}}_i(t_k)\|_2^2.
\end{equation}
where $M_{k,i} \in \{0,1\}$ indicates whether vertex $i$ 
is observed at frame $t_k$ and $\mathbf{x}_i(t_k) = \mathbf{X}_\phi(t_k, \mathbf{s}_i)$ 
is the network prediction at frame time $t_k$.
The overall objective is thus,
\begin{equation}
\min_{\phi,\theta}\quad
\lambda_{\text{phys}}\mathcal{L}_{\text{phys}}+
\lambda_{\text{bc}}\mathcal{L}_{\text{bc}}+
\lambda_{\text{ic}}\mathcal{L}_{\text{ic}}+
\lambda_{\text{data}}\mathcal{L}_{\text{data}}.
\end{equation}\label{eq:mainobj}
where $\lambda_{\text{phys}}, \lambda_{\text{bc}}, 
\lambda_{\text{ic}},$ and $\lambda_{\text{data}}$ 
are loss weights.
For forward simulation, $\theta$ is provided and fixed during training.

\section{Experiments}
\label{sec:experiment}
We evaluate our approach across physics-only and data-scarce settings. Hence, we assess ContinuousCloth-MLP and ContinuousCloth-Former, as pure physics-based ODE solvers (Sec.~\ref{sec:Physics-only Forward Simulation}), benchmarking numerical accuracy against classical time-stepping methods via trajectory error, comparing runtime efficiency, and examining robustness under diverse boundary and parameter conditions. Then, we evaluate data-scarce inference where measurements are sparse and noisy in different levels (clean, low noise and high noise)(Sec.~\ref{sec:Data-driven Interpolation and Parameter Estimation}), demonstrating (i) interpolation from limited or noisy frames and (ii) inverse estimation of physical parameters from motion, including experiments on real-world cloth sequences.

\subsection{Physics-only Forward Simulation}\label{sec:Physics-only Forward Simulation}

We evaluate the accuracy of ContinuousCloth in a physics-only setting by comparing it to a high-precision numerical solver. Because training uses stochastic mini-batches of time samples, low spring mass residual (training loss) alone does not guarantee uniformly accurate dynamics across the full time horizon. We therefore validate performance explicitly over the entire rollout interval.
\vspace{2mm}
\noindent\textbf{Reference solver and metrics.}
We use a fourth-order Runge--Kutta (RK4) \cite{yang2013cloth} integrator with a small time step (dt=$1\times10^{-4}$) as ground-truth. 
Under identical physical parameters and initial/boundary conditions, we report RMSE in meters and the mean spring-mass residual accumulated over the full trajectory.

\vspace{2mm}
\noindent\textbf{Physical model.}
We model the fabric as a rectangular spring--mass grid with resolution $n_x=16$ and $n_y=16$. Vertex masses are computed from a fixed material density $\rho=250$ and thickness $5\times10^{-4}\,\mathrm{m}$. We apply a constant gravity $\mathbf{g}=(0,0,-9.81)\,\mathrm{m/s^2}$. Structural, shear, and bending stiffness are fixed to $k_{\mathrm{\textit{struct}}}=1.0$, $k_{\mathrm{\textit{shear}}}=0.8$, and $k_{\mathrm{\textit{bend}}}=0.01$. Damping is modeled with per-spring coefficients $(c_{\mathrm{\textit{struct}}},c_{\mathrm{\textit{shear}}},c_{\mathrm{\textit{bend}}})=(0.005,0.003,0.002)$ and a global damping term $c_{\mathrm{\textit{global}}}=0.002$. 
All rollouts span $t \in [0,3]$ seconds. 

\vspace{2mm}
\noindent\textbf{Scenarios.}
We evaluate six scenarios spanning diverse boundaries and initial conditions. Scenarios S1–S4 use \emph{static (fixed)} boundary constraints with the cloth initialized flat on the $x$--$z$ plane and deformed solely by gravity. We vary the constrains by fixing a single corner vertex (S1), two diagonally opposite corner vertices (S2), three corner vertices (S3) and all corner vertices (S4). Scenarios S5-S6 use \emph{time-varying (moving)} boundary conditions with the cloth initialized on the $x$--$y$ plane. In \emph{lifting} (S5), two boundary corner vertices on the top edge move upward by $1.0\,\mathrm{m}$ over $t\in[0,2.0]$~s. In \emph{dynamic-right} (S6), the same boundry corner vertices translate by $1.0\,\mathrm{m}$ along the positive $x$-direction over $t\in[0,2.0]$~s. 
Illustrations for each scenario are provided in the supplementary material. 

\begin{wrapfigure}{r}{0.4\textwidth}
   \centering
   \includegraphics[width=\linewidth]{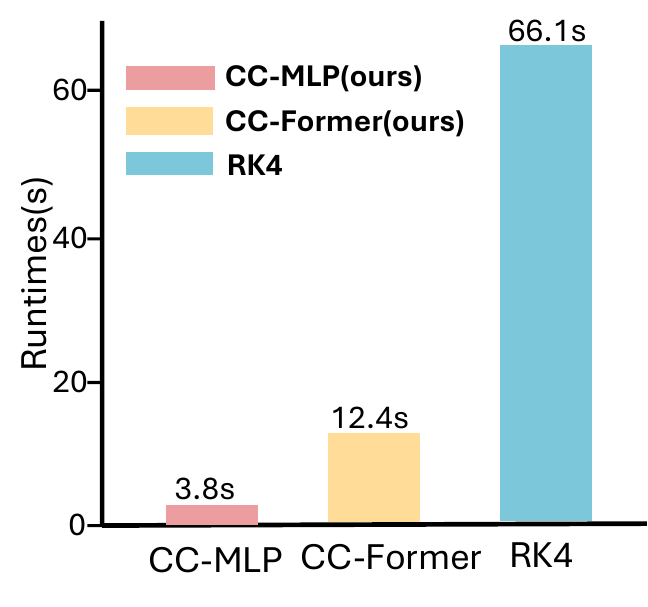}
   \caption{Average of per-rollout runtime across scenarios: RK4 (fixed step size) vs.\ our ContinuousCloth (CC) method.}
   \label{fig:time_comparison}
\end{wrapfigure}



\subsubsection{Accuracy and Efficiency.}

\begin{figure*}
    \center
    \includegraphics[width=1.0\linewidth]{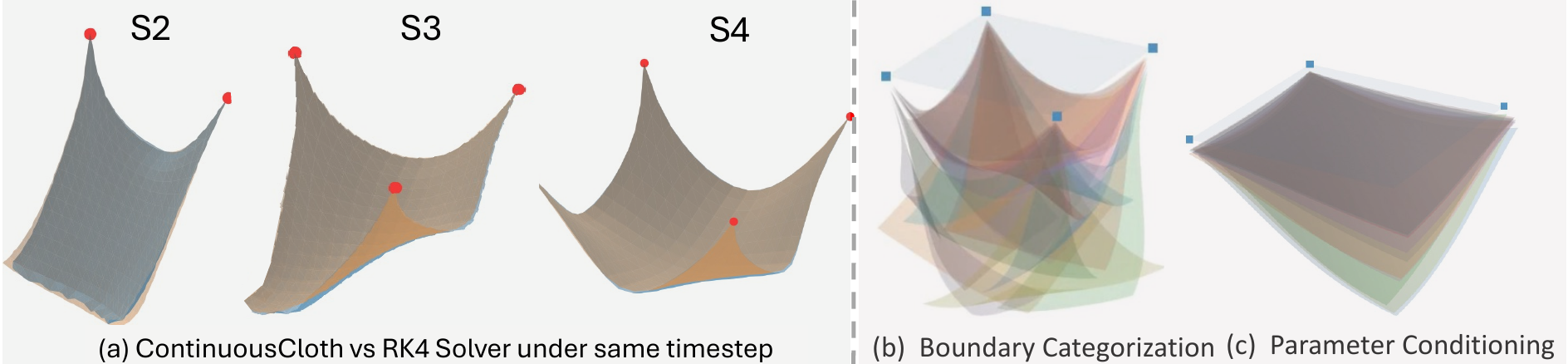}
    \vspace{3mm}
    \caption{Qualitative results in the physics-only forward simulation. (a) Trajectory overlay (blue: ground truth, orange: prediction) of ContinuousCloth solver and an RK4 reference\cite{yang2013cloth} under the same evaluation step size. Three scenarios are S2, S3 and S4. (b) Boundary-mask conditioning enables a single model to handle diverse constraint patterns. (c) Parameter conditioning allows the solver to generalize across different material parameters.}
    \label{fig:threeplot}
\end{figure*}
\begin{figure*}
    \centering
    \includegraphics[width=1.0\linewidth]{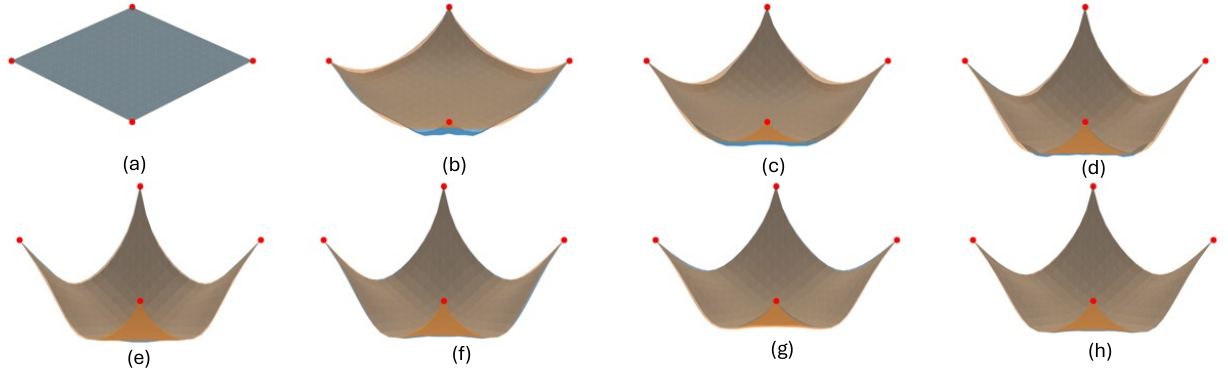}
    \vspace{3mm}
    \caption{
    Oscillatory cloth dynamics in the four-corner pinned scenario (S4). Blue shows the high-precision RK4 reference trajectory, and orange shows the ContinuousCloth prediction. The close alignment demonstrates that our model captures the underlying spring--mass oscillations over time.}
    \label{fig:planes}
\end{figure*}

For this, we implement two MLP-based PINN variants as baselines, \textit{QRes}\cite{bu2021quadratic} and \textit{FLS}\cite{wong2022learning}. As shown in Table~\ref{tab:scenario_compare}, both our method ContinuousCloth-MLP and ContinuousCloth-Former achieve consistently low physics residual (PR) and trajectory RMSE relative to the RK4 reference\cite{yang2013cloth} across all six scenarios, demonstrating that the learned dynamics remain accurate over the full continuous-time interval rather than only at sampled training points. Note that ContinuousCloth-MLP and ContinuousCloth-Former are trained in a data-free setting and learning is solely based on the physics residual. Therefore, ContinuousCloth-MLP attains a PR of $8.46\times10^{-4}$ N in S1 compared to $1.17\times10^{-3}$ N for QRes\cite{bu2021quadratic} and $4.37\times10^{-3}$ N for FLS\cite{wong2022learning}, and ContinuousCloth-Former achieves an RMSE of $3.19\times10^{-3}$ m in S4 against 
$3.76\times10^{-1}$ m for QRes\cite{bu2021quadratic}. Incorporating MFPSG consistently improves both metrics across all variants, with ContinuousCloth-MLP showing a $4.6\times$ reduction in PR and $9.1\times$ in RMSE when MFPSG is enabled which demonstrates the importance for capturing high-frequency cloth dynamics. Trajectory overlay visualizations (Fig.~\ref{fig:threeplot}-(a)) further confirm strong temporal alignment with the RK4 reference\cite{yang2013cloth} and no observable drift, qualitatively supporting the stability of our continuous-time formulation. Moreover, scenario(S4) demonstrates in Fig.~\ref{fig:planes} that the fabric undergoes oscillatory motion over time, suggesting that the model is not simply interpolating between observations or memorizing the final state, but is instead learning the true underlying dynamical ODE.

We also compare the average per-rollout runtime between RK4\cite{yang2013cloth} and our ContinuousCloth across six scenarios (See Fig. \ref{fig:time_comparison}). ContinuousCloth-MLP completes each rollout in $3.8\pm 0.07\,\mathrm{s}$ and ContinuousCloth-Former in $12.4\pm 0.11\mathrm{s}$, compared to $66.1\pm 0.06\mathrm{s}$ for RK4 at $\Delta t = 1\times10^{-4}$, yielding speedups of $17.4\times$ and $5.3\times$ respectively. Unlike RK4, whose cost scales linearly with the number of integration steps, ContinuousCloth evaluates the full trajectory in a single forward pass, decoupling inference cost from temporal resolution.


\FloatBarrier
\begin{table}[!htbp]
\centering
\caption{Quantitative comparison across S1–S6 scenarios. We report the mean physics residual (PR) in newtons and RMSE in meters over the full rollout horizon for each method. 'w/o MFPSG' denotes an ablated variant without MFPSG. 'CC' represents ContinuousCloth.}
\vspace{3mm}
\label{tab:scenario_compare}
\resizebox{1.0\linewidth}{!}{%
\begin{tabular}{l *{6}{ccc}}
\toprule
\multirow{2}{*}{Method} 
  & \multicolumn{2}{c}{S1} 
  & \multicolumn{2}{c}{S2} 
  & \multicolumn{2}{c}{S3} 
  & \multicolumn{2}{c}{S4}
& \multicolumn{2}{c}{S5}
& \multicolumn{2}{c}{S6}\\
\cmidrule(lr){2-3} \cmidrule(lr){4-5} \cmidrule(lr){6-7} \cmidrule(lr){8-9}\cmidrule(lr){10-11}\cmidrule(lr){12-13}
  & {PR $\downarrow$} & {RMSE $\downarrow$}
  & {PR $\downarrow$} & {RMSE $\downarrow$}
  & {PR $\downarrow$} & {RMSE $\downarrow$}
  & {PR $\downarrow$} & {RMSE $\downarrow$}
   & {PR $\downarrow$} & {RMSE $\downarrow$} 
    & {PR $\downarrow$} & {RMSE $\downarrow$} \\
\midrule
QRes(w/o MFPSG)        
  & $4.39\times10^{-3}$ & $7.67\times10^{-1}$
  & $4.88\times10^{-3}$ & $5.72\times10^{-1}$
  & $8.13\times10^{-2}$ & $4.65\times10^{-1}$
  & $9.31\times10^{-2}$ & $3.75\times10^{-1}$ 
   & $4.92\times10^{-2}$ & $7.06\times10^{-1}$
  & $4.91\times10^{-2}$ & $7.01\times10^{-1}$\\
QRes       
  & $1.17\times10^{-3}$ & $1.24\times10^{-1}$
  & $1.73\times10^{-3}$ & $1.01\times10^{-1}$
  & $7.98\times10^{-2}$ & $4.66\times10^{-1}$
  & $9.17\times10^{-2}$ & $3.76\times10^{-1}$
  & $3.93\times10^{-2}$ & $6.44\times10^{-1}$
  & $4.11\times10^{-2}$ & $6.67\times10^{-1}$\\
FLS(w/o MFPSG)         
  & $4.51\times10^{-3}$ & $7.93\times10^{-1}$
  & $4.68\times10^{-3}$ & $5.55\times10^{-1}$ 
  & $4.86\times10^{-3}$ & $3.90\times10^{-1}$
  & $4.95\times10^{-3}$ & $2.73\times10^{-1}$
  & $5.01\times10^{-3}$ & $4.72\times10^{-1}$
  & $5.15\times10^{-3}$ & $5.41\times10^{-1}$\\
FLS         
  & $4.37\times10^{-3}$ & $6.68\times10^{-1}$
  & $4.53\times10^{-3}$ & $4.39\times10^{-1}$ 
  & $4.77\times10^{-3}$ & $3.64\times10^{-1}$
  & $4.87\times10^{-3}$ & $2.55\times10^{-1}$ 
   & $4.76\times10^{-3}$ & $3.84\times10^{-1}$
  & $5.92\times10^{-3}$ & $5.34\times10^{-1}$ \\
CC-MLP(w/o MFPSG)         
  & $2.02\times10^{-3}$ & $1.19\times10^{-1}$
  & $1.59\times10^{-3}$ & $1.01\times10^{-1}$ 
  & $2.57\times10^{-3}$ & $9.58\times10^{-2}$  
  & $3.21\times10^{-3}$ & $1.12\times10^{-1}$
   & $2.55\times10^{-2}$ & $8.19\times10^{-1}$
  & $2.27\times10^{-2}$ & $7.30\times10^{-1}$\\
CC-MLP (ours)        
  & \bm{$8.46\times10^{-4}$} & $9.11\times10^{-2}$ 
  & $6.88\times10^{-4}$ & $1.44\times10^{-2}$ 
  & \bm{$6.85\times10^{-4}$} & \bm{$1.69\times10^{-2}$}
  & \bm{$7.05\times10^{-4}$} & $6.90\times10^{-3}$
   & \bm{$4.15\times10^{-3}$} & \bm{$3.62\times10^{-2}$}
  & \bm{$5.39\times10^{-3}$} & \bm{$5.24\times10^{-2}$}\\

PINNsFormer(w/o MFPSG)
  & $1.56\times10^{-3}$ & $1.16\times10^{-1}$ 
  & $4.79\times10^{-3}$ & $6.05\times10^{-1}$ 
  & $4.88\times10^{-3}$ & $4.14\times10^{-1}$
  & $5.08\times10^{-3}$ & $2.89\times10^{-1}$ 
  & $1.64\times10^{-2}$ & $7.57\times10^{-1}$
  & $1.09\times10^{-2}$ & $6.88\times10^{-1}$\\
CC-Former (ours)     
  & $9.69\times10^{-4}$ & \bm{$8.99\times10^{-2}$} 
  & \bm{$4.61\times10^{-4}$} & \bm{$1.20\times10^{-2}$} 
  & $6.91\times10^{-4}$ & $1.75\times10^{-2}$ 
  & $1.07\times10^{-3}$ & \bm{$3.19\times10^{-3}$}
   & $7.49\times10^{-3}$ & $6.66\times10^{-2}$
  & $9.07\times10^{-3}$ & $6.83\times10^{-2}$\\

\bottomrule  
\end{tabular}%
}
\end{table}
\FloatBarrier


\subsubsection{Cross-Condition Generalization}

Beyond fitting a single dynamical scenario, we aim to learn a \emph{scenarios-level} cloth dynamics model that generalizes across both \emph{boundary constraints} and \emph{material parameters}. 
To support diverse constraint patterns within a unified model, we condition on a binary vertex mask that specifies constrained vertices while keeping the initial state fixed. In total, we jointly train on 15 distinct boundary-condition settings within one unified network. 
We further condition the network on a physical parameter vector (e.g., stiffness and damping) and train with parameter resampling. That is, each iteration samples a new parameter set and optimizes the model using a mini-batch of $1{,}000$ time samples. Quantitatively, ContinuousCloth-MLP achieves spring--mass residuals of $7.32\times10^{-4}$ N and $4.73\times10^{-4}$ N with RMSEs of $3.37\times10^{-2}$ m and $8.25\times10^{-2}$ m under boundary-condition and parameter generalization, respectively.
Qualitative rollouts (Fig.~\ref{fig:threeplot}-(b) and (c)) demonstrate temporally continuous trajectories with stable, smooth oscillatory behavior across both parameter variations and boundary masks.

\subsection{Data-scarce Interpolation and Parameter Estimation Tasks}\label{sec:Data-driven Interpolation and Parameter Estimation}
We evaluate our method in data-scarce settings with sparse and noisy measurements, considering both simulated and real-world fabric trajectories. By combining observations with the spring--mass ODE constraint, we address two tasks: (i) interpolation of continuous cloth trajectories from limited frames, and (ii) inverse estimation of physical parameters from motion. As ContinuousCloth-MLP and ContinuousCloth-Former yield comparable accuracy (Sec.~\ref{sec:Physics-only Forward Simulation}) while ContinuousCloth-MLP is faster, we adopt ContinuousCloth-MLP for all experiments in this section. We compare against a data-only MLP baseline to isolate the contribution of physics guidance.

\vspace{2mm}
\noindent\textbf{Simulation Data Collection.}
We generate synthetic fabric sequences using the spring-mass solver fourth-order Runge--Kutta (RK4)\cite{yang2013cloth} with fixed physical parameters. Each rollout simulates a $1\,\mathrm{m}\times1\,\mathrm{m}$ cloth over $T=3$~s and is uniformly subsampled into $10$ frames per second. 
We consider three boundary scenarios: (D1) Bouncing (two pinned corner vertices with the cloth initialized vertically); (D2) Moving (moving boundary, where the pinned vertices translate $1$\,m horizontally over $t \in [0, 2]$\,s); and (D3) Swing (two pinned vertices with the cloth initialized horizontally).

To simulate sensor noise, we generate three observation variants per scenario: \textit{clean}, \textit{low-noise}, and \textit{high-noise}, obtained by injecting Gaussian jitter, drift, and sparse outliers into the simulated trajectories, while pinned vertices remain noise-free. The clean and noise settings use different physical parameter 
configurations. That is, the clean setting adopts stiffer parameters to produce highly nonlinear dynamics, testing the model's ability to capture rapid oscillations; the noisy settings use softer parameters to produce smoother trajectories, shifting the challenge towards robustness under noisy observations. Visualization of data collection scenarios and parameter details are provided in the supplementary material.

\vspace{1mm}
\noindent\textbf{Real-world Data Collection.}
We collect a real-world fabric dataset with three fabrics of identical size ($35\,\mathrm{cm}\times 35\,\mathrm{cm}$) but distinct physical behavior, ranging from highly deformable to relatively stiff: \emph{camel ponte roma}, \emph{black denim}, and \emph{red jet set}\cite{wang2011data}. 
We record synchronized RGB-D streams using a front-mounted ZED2i camera at $30$~Hz while a dual-arm robot executes predefined manipulation trajectories in zero-G mode~\cite{armcontrolsystem}. End-effector poses are recorded at $30$~Hz and time-aligned with RGB-D frames. We evaluate two scenarios:
(R1) Shaking (one top corner fixed, one top corner moving) and (R2) Fling (both top corners moving) .
We obtain 3D cloth observations by tracking surface markers through the RGB-D sequence, details are provided in the supplementary material.
In our recordings, we track 90 points over 50 frames per scenario.
To apply the spring--mass model, we estimate a spring topology from the first frame and build structural, shear, and bending springs based on local neighborhood relations.

\begin{table}[!htbp]
\centering
\caption{Interpolation RMSE (in meters) on simulated cloth trajectories (lower is better) across three scenarios (D1: Bouncing, D2: Moving, D3: Swing) and three observation conditions (clean/low-noise/large-noise). CC-MLP represent ContinuousCloth-MLP. }
\vspace{4mm}
\label{tab:inter_sim}
\begin{adjustbox}{max width=0.95\linewidth}
\begin{tabular}{l c c c  c c c  c c c}
\toprule
\multirow{3}{*}{Method} &
\multicolumn{3}{c}{D1 Bouncing} &
\multicolumn{3}{c}{D2 Moving} &
\multicolumn{3}{c}{D3 Swing} \\
\cmidrule(lr){2-4}\cmidrule(lr){5-7}\cmidrule(lr){8-10}
& Clean & Low noise & High noise
& Clean & Low noise & High noise
& Clean & Low noise & High noise \\
\midrule
MLP
& $1.48\times10^{-1}$ & $3.16\times10^{-1}$ & $4.60\times10^{-1}$
& $1.24\times10^{-1}$ & $2.78\times10^{-1}$ & $4.32\times10^{-1}$
& $8.03\times10^{-2}$ & $3.05\times10^{-1}$ & $4.27\times10^{-1}$ \\
CC-MLP
& $\bm{3.72\times10^{-2}}$ & $\bm{4.74\times10^{-2}}$ & $\bm{6.77\times10^{-2}}$
& $\bm{2.09\times10^{-2}}$ & $\bm{3.78\times10^{-2}}$ & $\bm{5.93\times10^{-2}}$
& $\bm{2.63\times10^{-2}}$ & $\bm{2.23\times10^{-2}}$ & $\bm{3.26\times10^{-2}}$ \\
\bottomrule
\end{tabular}
\end{adjustbox}
\end{table}

\subsubsection{Trajectory Interpolation}


\begin{wraptable}{r}{0.65\textwidth}
\centering
\caption{Interpolation RMSE (in meters) on real-world cloth sequences (lower is better) for R1 (Shaking) and R2 (Fling) across three fabrics: Camel ponte roma (Camel), Black denim (Black), and Red jet set (Red). CC-MLP represents ContinuousCloth-MLP.}
\vspace{3mm}
\label{tab:inter_real}
\begin{adjustbox}{max width=1\linewidth}
\begin{tabular}{l c c c  c c c}
\toprule
\multirow{2}{*}{Method} &
\multicolumn{3}{c}{R1 Shaking} &
\multicolumn{3}{c}{R2 Fling} \\
\cmidrule(lr){2-4}\cmidrule(lr){5-7}
& Camel  & Black& Red 
& Camel & Black  & Red
 \\
\midrule
MLP
& $1.33\times10^{-2}$ & $9.79\times10^{-3}$ & $9.68\times10^{-3}$
& $3.76\times10^{-2}$ & $3.05\times10^{-2}$ & $4.318\times10^{-2}$ \\
CC-MLP
& $\bm{7.39\times10^{-3}}$ & $\bm{5.94\times10^{-3}}$ & $\bm{6.19\times10^{-3}}$
& $\bm{1.38\times10^{-2}}$ & $\bm{2.09\times10^{-2}}$ & $\bm{5.93\times10^{-3}}$ \\
\bottomrule
\end{tabular}
\end{adjustbox}
\end{wraptable}

In both simulated and real-world settings, we evaluate interpolation by training on every 10th frame and reconstructing all intermediate frames. We train with a combined objective of data fitting, boundary constraints, 
initial conditions, and physics residuals (Eq. \ref{eq:mainobj}). The data-only baseline removes 
the physics residual while retaining all other terms. Performance is evaluated via RMSE between predicted and ground-truth 3D 
positions over all timestamps in meters; when training observations are noisy, evaluation is still performed against the clean ground-truth trajectory. The clean simulated setting uses the same physical parameters as Sec.~\ref{sec:Physics-only Forward Simulation}. For real-world sequences, fabric parameters are unknown, we, therefore, first estimate them via inverse PINN fitting and then perform interpolation using the identified parameters.

\begin{figure*}[t]
    \center
    \includegraphics[width=0.85\linewidth]{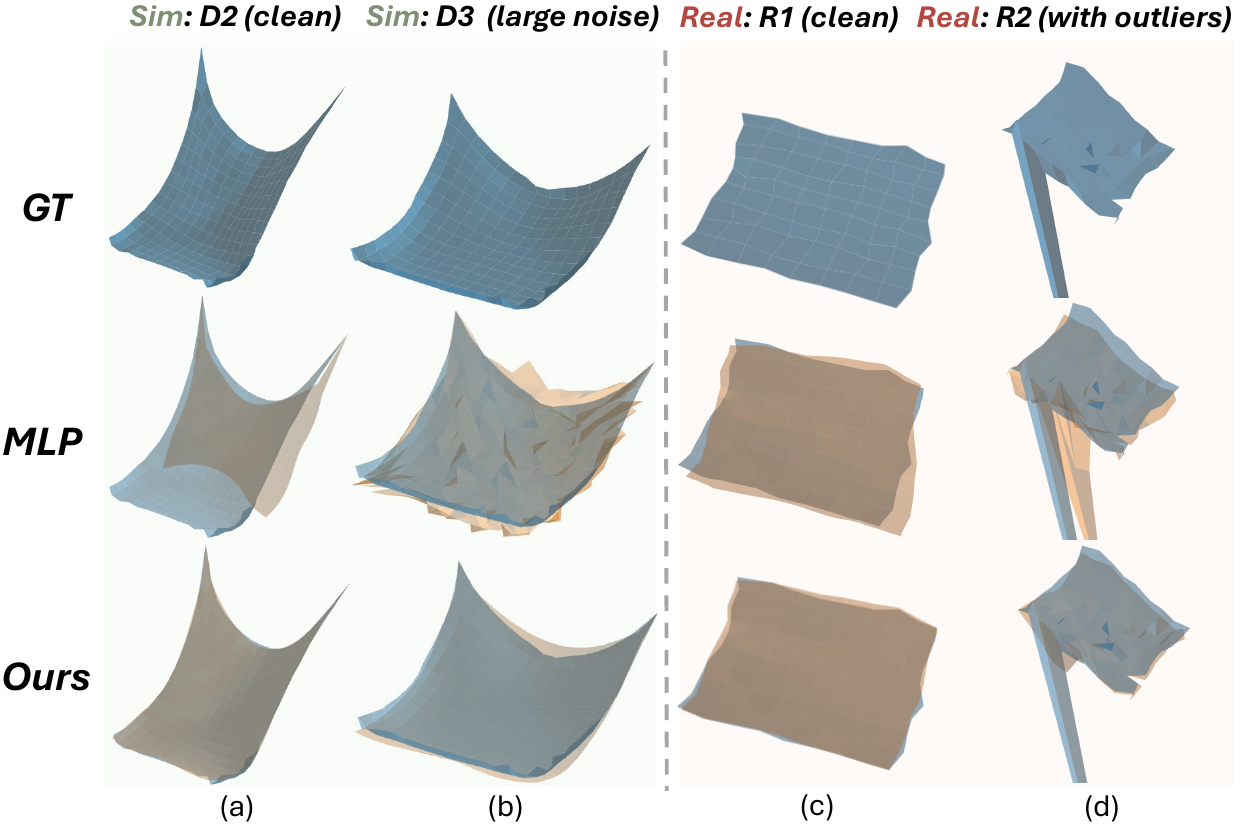}
    \vspace{3mm}
    \caption{Qualitative interpolation results across simulation (D2 Moving clean, D3 Swing large-noise) and real data (R1 Shaking clean, R2 Fling outliers). Compared with the MLP baseline, our ContinuousCloth yields more stable and physically plausible trajectories, especially under noisy observations.}
    \label{fig:visulize}
    \vspace{-3mm}
\end{figure*}

As shown in Tables~\ref{tab:inter_sim} and~\ref{tab:inter_real}, 
ContinuousCloth-MLP consistently outperforms the data-only MLP across all scenarios and noise levels in both simulation and real world. In simulation, our method reduces the average RMSE across scenarios by $4.3\times$ in the clean setting and $8.3\times$ under high-noise. In the real world, it achieves up to $7.3\times$ lower RMSE on the red jet set fabric in R2 Fling ($5.93\times10^{-3}$ m vs.\ $4.32\times10^{-2}$ m). Fig.~\ref{fig:visulize} further supports these results. Our method remains stable under increasing noise while MLP accuracy drops significantly (Fig.~\ref{fig:visulize}(b)), captures high-frequency dynamics missed by the MLP in simulation (Fig.~\ref{fig:visulize}(a)), produces more consistent interpolations matching observed deformations (Fig.~\ref{fig:visulize}(c)), and is robust to tracking outliers and spikes in real-world sequences (Fig.~\ref{fig:visulize}(d)). 
Overall, our ContinuousCloth demonstrates strong interpolation performance across both simulated and real-world conditions, confirming the value of physics guidance for robust and accurate trajectory reconstruction.



%

\subsubsection{Physical Parameter Estimation}


We evaluate inverse parameter estimation in both simulated and real-world settings. We compare against a differentiable RK4 baseline, implemented 
by unrolling integration steps and backpropagating through the solver via automatic and Diffcloud\cite{sundaresan2022diffcloud}. In simulation, where ground-truth parameters are known, we estimate structural stiffness $k_{\mathrm{struct}}$ across different boundary conditions and noise levels, using the same physical setup as Sec.~\ref{sec:Physics-only Forward Simulation}. For real-world evaluation, we adopt a cross-condition from~\cite{ru2025can}. That is, models are trained on the flinging scenario (R2) to identify fabric parameters, which are then transferred to the shaking scenario (R1) and evaluated via held-out RMSE.
Parameters are optimized by combining physics residuals at randomly sampled timestamps with data terms at observed frames.

\begin{wraptable}{r}{0.50\textwidth}
\centering
\caption{Held-out rollout RMSE (in meters) for real-world parameter estimation on R1 (Shaking) cross three fabrics: Camel ponte roma (Camel), Black denim (Black), and Red jet set (Red). DS: differentiable-solver RK4; Diffcloud: a differentiable elastodynamic simulator-based real-to-sim pipeline; CC-MLP: ContinuousCloth-MLP.}
\label{tab:pp_real}
\begin{adjustbox}{max width=\linewidth}
\begin{tabular}{l c c c  c c c}
\toprule
\multirow{2}{*}{Method} &
\multicolumn{3}{c}{R1 Shaking}\\
\cmidrule(lr){2-4}\cmidrule(lr){5-7}
& Camel & Black & Red \\
\midrule
DS & $3.86\times10^{-2}$ & $3.42\times10^{-2}$ & $7.24\times10^{-2}$ \\
Diffcloud & \bm{$3.71\times10^{-2}$} &
$2.90\times10^{-2}$ &
$7.66\times10^{-2}$
\\ 
CC-MLP                & {$3.76\times10^{-2}$} & \bm{$2.32\times10^{-2}$} & \bm{$5.70\times10^{-2}$} \\
\bottomrule
\end{tabular}
\end{adjustbox}
\end{wraptable}

As summarized in Table~\ref{tab:pp_sim}, ContinuousCloth-MLP and the 
differentiable-solver RK4\cite{yang2013cloth} achieve comparable accuracy in the noise-free setting in simulation. Under noisy observations, ContinuousCloth-MLP is significantly more robust as the averaged over all scenarios in the high-noise setting, our method achieves a relative parameter error of $2.41\times10^{-1}$  compared to $6.16\times10^{-1}$  for the differentiable-solver RK4\cite{yang2013cloth}. We attribute this to the fact that simulator-based optimization fits parameters directly to noisy measurements, causing the optimizer to absorb  noise-induced observations into the parameter estimates. Furthermore, velocity and acceleration are approximated via finite differences over consecutive noisy frames, where the noise contribution is amplified by the inverse of the timestep, $\Delta t$, destabilizing gradient-based parameter updates. In contrast, ContinuousCloth-MLP enforces the ODE at randomly sampled collocation points across the full time horizon, decoupling the physics constraint from any fixed noisy discretization. For the real-world parameter estimation task, as shown in Table~\ref{tab:pp_real}, 
ContinuousCloth-MLP achieves lower held-out RMSE than the differentiable RK4 solver~\cite{yang2013cloth} across all three fabrics, with the largest improvement on the red jet set fabric 
(5.70 cm vs.7.24 cm). Compared with DiffCloud, a differentiable elastodynamic real-to-sim pipeline, ContinuousCloth-MLP achieves lower errors on black denim (2.32 cm vs. 2.90 cm) and red jet set (5.70 cm vs. 7.66 cm). Although DiffCloud uses a more plausible physically simulator for complex cloth dynamics, sparse trajectory observations make the inverse problem under-constrained because multiple parameter settings can explain the measured motion similarly well. Therefore, this expressiveness is difficult to identify reliably and may not translate into better cross-scenario accuracy, as shown in Table~\ref{tab:pp_real} and consistent with previous findings on diffcloud limitations on real2sim\cite{ru2025can}.


\begin{table}[!t]
\centering
\caption{Parameter estimation performance in simulation (relative error). We compare a differentiable-solver RK4 (DS) and ContinuousCloth-MLP (CC-MLP) across three simulated scenarios (D1 Bouncing, D2 Moving, D3 Swing) under clean, low-noise, and large-noise observations.}
\vspace{4mm}
\label{tab:pp_sim}
\begin{adjustbox}{max width=\linewidth}
\begin{tabular}{l c c c  c c c  c c c }
\toprule
\multirow{2}{*}{Method} &
\multicolumn{3}{c}{D1 Bouncing} &
\multicolumn{3}{c}{D2 Moving} &
\multicolumn{3}{c}{D3 Swing} \\
\cmidrule(lr){2-4}\cmidrule(lr){5-7}\cmidrule(lr){8-10}
& Clean & Low noise & High noise
& Clean & Low noise & High noise
& Clean & Low noise & High noise\\
\midrule
DS
& $\bm{9.15\times10^{-2}}$ & $6.57\times10^{-1}$ & $6.01\times10^{-1}$
& $7.82\times10^{-2}$ & $3.08\times10^{-1}$ & $4.27\times10^{-1}$
& $\bm{4.72\times10^{-2}}$ & $2.85\times10^{-1}$ & $8.19\times10^{-1}$ \\
CC-MLP
& $9.82\times10^{-2}$ & $\bm{3.51\times10^{-1}}$ & $\bm{3.91\times10^{-1}}$
& $\bm{7.40\times10^{-2}}$ & $\bm{9.95\times10^{-2}}$ & $\bm{1.60\times10^{-1}}$
& $5.23\times10^{-2}$ & $\bm{0.58\times10^{-1}}$ & $\bm{1.72\times10^{-1}}$ \\
\bottomrule
\end{tabular}
\end{adjustbox}
\end{table}

\vspace{-3mm}
\section{Conclusion}

In this work, we presented \textbf{ContinuousCloth}, a continuous-time physics-informed neural solver for deformable cloth dynamics based on the spring--mass ODE equation. By representing cloth motion as a differentiable trajectory field, ContinuousCloth decouples trajectory evaluation from a fixed integration grid and enables direct queries of positions and derivatives at arbitrary timestamps. We implemented our method with both MLP and transformer backbones, showing that the proposed continuous-time framework is architecture-agnostic.

Our experiments suggest two main insights. First, for physics-only forward solving, a continuous neural trajectory field can approximate high-precision time-stepping dynamics with strong numerical fidelity while substantially reducing dense-time evaluation cost by up to 17.4x. Second, in data-scarce settings, the spring--mass ODE provides an effective physical prior where full-horizon residual regularization improves interpolation from sparse observations and makes inverse parameter estimation and interpolation more robust to measurement noise. These results indicate that continuous-time physics-informed cloth modeling is particularly useful when cloth trajectories must be queried densely, differentiated accurately, or facing sparse and noisy observations, for applications such as generative cloth state estimation under partial observability\cite{tian2025diffusion} and real2sim parameter estimation task\cite{sundaresan2022diffcloud}.

Overall, ContinuousCloth shows that it is a unified framework which includes forward simulation, interpolation and parameter estimation. Although our current framework does not yet model contact-rich scenarios such as collision, self-contact, and friction, extending continuous-time physics-informed cloth models to these settings is an interesting direction for future work.




%
%
\bibliographystyle{splncs04}
\bibliography{egbib}
\end{document}